\documentclass[letterpaper]{article} 
\usepackage[preprint]{aaai2027}  
\usepackage[hyphens]{url}  
\usepackage{graphicx} 
\usepackage{natbib}  
\usepackage{caption} 
\usepackage{algorithm}
\usepackage{algorithmic}
\usepackage{amsmath}
\usepackage{amssymb}
\usepackage{svg}
\usepackage{newfloat}
\usepackage{listings}
\DeclareCaptionStyle{ruled}{labelfont=normalfont,labelsep=colon,strut=off} 
\floatstyle{ruled}
\newfloat{listing}{tb}{lst}{}
\floatname{listing}{Listing}

\usepackage{booktabs}

\usepackage{tabularx}

\usepackage{diagbox}

\usepackage{enumitem}

\title{EEG-AS: Instance-Level Foundation Model Selection for EEG Foundation Models via Behavior Reconstruction}

\author{
    Yunzhen Zhang\textsuperscript{\rm 1}\equalcontrib,
    Ruoxi Piao\textsuperscript{\rm 1}\equalcontrib,
    Muhammad Ibrahim Ali Shah\textsuperscript{\rm1},
    Hasan Onur Keles\textsuperscript{\rm 2},
    Mustafa Misir\textsuperscript{\rm 1}\corresponding
}

\affiliations{
    \textsuperscript{\rm 1}Duke Kunshan University\\
    \textsuperscript{\rm 2}Ankara University
}

\begin{document}

\maketitle

\begin{abstract}
Electroencephalography (EEG) is a non-invasive technique for measuring neural activity and has been widely used in neuroscience applications. Recent advances in EEG foundation models have enabled strong performance across diverse neural decoding tasks. However, no single foundation model consistently performs best across datasets or individual EEG instances, while instance-level model selection remains largely unexplored. To address this limitation, we formulate EEG foundation model selection as an instance-level Algorithm Selection (AS) problem. We propose \textbf{EEG-AS}, an instance-level algorithm selection framework that characterizes each EEG instance using inference-available latent EEG embeddings, handcrafted neurophysiological features, and an anchor foundation model. During training, EEG-AS learns to reconstruct unavailable foundation-model behaviors from privileged prediction tokens conditioned on an anchor foundation model, while during inference it estimates these behaviors without executing the entire model portfolio, enabling efficient selection from seven EEG foundation models. Experiments on seven public EEG benchmarks demonstrate that EEG-AS substantially narrows the gap between the Single Best Solver (SBS) and the oracle upper bound for each instance. These results highlight the effectiveness of instance-level AS for adaptive deployment of EEG foundation models.
\end{abstract}


\section{Introduction}
\label{sec:intro}
Electroencephalography (EEG) foundation models (FMs) have recently emerged as a promising paradigm for learning transferable neural representations from large-scale unlabeled EEG data \cite{kuruppu2026eegFoundationREV}. General-purpose models have demonstrated strong performance across a wide range of downstream EEG applications\cite{xiong2026eeg}. As the number of available EEG FMs continues to grow, \textbf{attention is shifting from developing stronger individual models to effectively utilizing existing ones.}

Current practice typically deploys a single FM according to its average performance on a benchmark dataset. This \textbf{Single Best Solver (SBS)} strategy is implicitly based on the assumption that one model is uniformly preferable for all EEG instances within a dataset. However, our empirical analysis reveals that this assumption does not hold. Although a FM may achieve the highest average accuracy, its performance varies substantially across individual EEG instances. Different FMs correctly classify different instances, \textbf{revealing substantial instance-level complementarity beyond their dataset-level competition.} This observation is further supported by the large performance gap between the SBS and the \textbf{Virtual Best Solver (VBS)} / \textbf{Oracle upper bound} observed across all evaluated datasets, suggesting promising potential for adaptive model deployment. These observations echo the common \textbf{Algorithm Selection (AS)} premise that algorithm performance is instance-dependent \cite{kerschke2019automated}. Since EEG inference is ultimately performed on individual trials, windows, or recordings, this heterogeneity motivates adapting model deployment at the same granularity.

These observations raise a fundamental research question: \textbf{Can we automatically identify the most suitable FM for each EEG instance?} We cast this problem as a per-instance AS task, where one FM is selected from a portfolio for each EEG instance \cite{kerschke2019automated}. 
We propose \textbf{EEG-AS}. Our key insight is that the behaviors of unexecuted candidate FMs can be inferred from an inference-available EEG representation together with the prediction behavior of a single inference-available anchor FM. During training, EEG-AS learns the relationship among different FMs using privileged prediction tokens generated offline. During inference, it reconstructs the unavailable model behaviors conditioned on the anchor model and selects the most suitable FM without exhaustively evaluating the entire portfolio. Experiments on seven public EEG benchmarks demonstrate that EEG-AS achieves the highest average performance among deployable methods, obtaining the best result on five datasets while remaining competitive on the others and outperforming SBS on all seven datasets.

The main contributions of this work are summarized as:
\begin{itemize}[leftmargin=*]
    \item \textbf{We formulate EEG FM deployment as an instance-level AS problem.} To the best of our knowledge, this is the first work on adaptive deployment of EEG FMs at the instance level, moving beyond the conventional single-model deployment. Two traditional AS models are built.
    \item \textbf{We propose EEG-AS, a novel instance-level selection framework that reconstructs unavailable FM behaviors from an EEG representation and a single inference-available anchor model.} This enables effective model selection without exhaustively executing the entire FM portfolio during inference. 
    \item \textbf{We conduct comprehensive experiments on seven public EEG benchmarks covering diverse downstream applications.} The experimental results demonstrate that EEG-AS improves average selection accuracy from 53.7\% for SBS to 66.9\%, closes 44.0\% of the SBS-to-Oracle performance gap and the 83.7\% Oracle upper bound. It also provides extensive analyses of instance-level complementarity among EEG FMs, highlighting the potential of AS for adaptive EEG FM deployment.
\end{itemize}

\section{Background}
\label{sec:related}

\subsection{EEG Foundation Models}
Recent EEG foundation models (FMs) leverage self-supervised pretraining on large-scale unlabeled EEG data to learn transferable neural representations for downstream EEG applications. Existing EEG FMs differ substantially in architectural design and pretraining strategy, reflecting diverse approaches to modeling the spatial, temporal, and semantic characteristics of EEG signals. Comprehensive benchmarks such as EEG-FM-Bench \cite{xiong2025eeg,xiong2026eeg} have standardized the evaluation of these models across diverse EEG datasets, providing a unified benchmark for comparing their downstream performance.

\subsection{Automated Algorithm Selection (AS)}


AS has been applied to various problem domains such as Boolean satisfiability, combinatorial optimization and machine learning. 
Within the scope of Automated Machine Learning (AutoML), AS focuses on a critical meta-learning task. 
Traditionally, AS is implemented through performance prediction models that identify the best expected candidate algorithm for a given problem instance. 
Many studies have focused on utilizing handcrafted features to build these models. 
To eliminate the need of domain expertise for feature engineering, representation learning has been investigated. 
Convolutional Neural Networks (CNNs) \cite{loreggia2016deep} is an early example in this direction. 
Following that, considering the structural nature of the target problems, Graph Neural Networks (GNNs) have been adopted to learn instance characterizations from graph data \cite{cao2025mcGNNASDock}. 
More recently, advances in Transformer architectures and large-scale embedding models have enabled the automatic extraction of high-dimensional latent representations \cite{wang2026molecular}.

\section{Method}
\label{sec:method}

\subsection{Problem Formulation}

Following the standard Automated Algorithm Selection (AS) formulation \cite{rice1976algorithm,kerschke2019automated},
we specify EEG Foundation Model (FM) selection as an per-instance AS problem over a portfolio of pretrained FMs.
Let
\(
\mathcal{M}=\{m_1,\ldots,m_K\}
\)
denote a portfolio of $K$ frozen EEG FMs
Although all models are evaluated on the same downstream EEG task within each dataset, their performances may vary substantially across different EEG instances due to their distinct architectures and learned representations.

For an EEG instance $x_i$ and a FM $m_k\in\mathcal{M}$, let
\(
\rho(m_k,x_i)
\)
is the utility of model $m_k$ on instance $x_i$. 
As the performance upper bound, Virtual Best Solver (VBS) / Oracle represents the theoretical performance limit, i.e. choosing the absolute best FM for each individual instance \cite{kerschke2019automated}:
\begin{equation}
m^*(x_i)
=
\arg\max_{m_k\in\mathcal{M}}
\rho(m_k,x_i).
\end{equation}

The goal is to learn a selector that predicts the best FM using only information available during inference. 
Specifically, the selector produces a relevance score $s(x_i,m_k)=f_{\theta}(x_i,m_k)$ for each FM.

and chooses the one with the highest score:
\begin{equation}
\hat m(x_i)
=
\arg\max_{m_k\in\mathcal{M}}s(x_i,m_k).
\end{equation}

The objective is to predict the Oracle model assignment $m^*(x_i)$ as accurately as possible without evaluating the entire FM portfolio for each EEG instance.
At the same time, the Single Best Solver (SBS) is required to be outperformed.
In this work, SBS is one FM delivering the best overall performance across the entire instance set.
During training, additional information from FMs may be utilized to learn model suitability, while inference-time selection only relies on information available before executing the candidate FM.

\subsection{EEG-AS: Overall Framework}
\begin{figure*}[t]
    \centering
    \includegraphics[width=\textwidth]{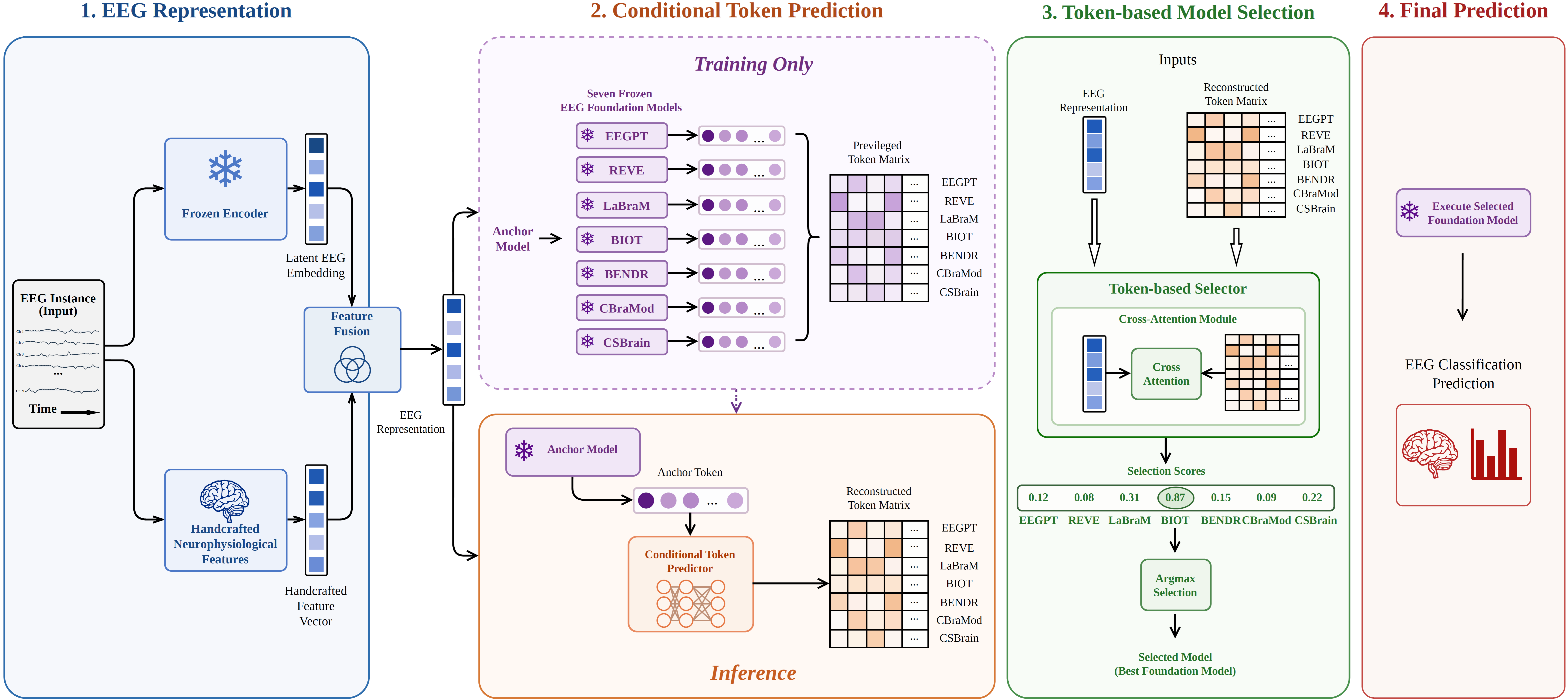}
    \caption{
    Overview of the EEG-AS.
    During training, privileged prediction tokens are available.
    During inference, only the anchor model is executed.
    }
    \label{fig:framework}
\end{figure*}

EEG-AS aims to perform efficient instance-level FM selection by reconstructing unavailable FM behaviors from a lightweight anchor model.
As shown in Fig.~\ref{fig:framework}, EEG-AS has three components: an EEG representation module, a conditional token predictor, and a token-based model selector.

Given an EEG instance, EEG-AS first extracts an inference-available representation. During training, prediction tokens from all FMs are available as privileged information. During inference, only the token from an anchor model is observed. The conditional token predictor reconstructs the missing FM tokens, and the reconstructed token matrix is passed to the selector to predict the most suitable FM for the current instance. The framework is optimized by combining token reconstruction and model selection objectives.



\subsubsection{EEG Representation.}
To characterize EEG instances without evaluating the entire FM portfolio, EEG-AS relies on an inference-available representation $z_x$. The framework is agnostic to the choice of representation, provided that it can be extracted without executing the candidate FMs. Such representations may be obtained from pretrained EEG encoders, handcrafted EEG features, or a combination of both.

Let $e_x=f_{\mathrm{emb}}(x)\in\mathbb{R}^{d_e}$
is the embedding extracted by an inference-time EEG encoder, and 
$h_x=f_{\mathrm{hc}}(x)\in\mathbb{R}^{d_h}
$
denote handcrafted EEG features describing complementary signal characteristics. The final EEG representation is obtained by
$
z_x=[e_x;h_x]
$.
In this work, $f_{\mathrm{emb}}$ is instantiated using a frozen BIOT encoder \cite{yang2023biot}.

\subsubsection{Privileged Token Representation.}
During training, EEG-AS has access to the prediction outputs of all foundation models (FMs) as privileged information. \cite{vapnik2009new} Rather than using scalar performance scores, we represent the behavior of each FM by its prediction token.

For an EEG instance $x$, the $k$-th FM produces a prediction token $t_k=\phi_k(x)\in\mathbb{R}^{C}$ where $\phi_k(\cdot)$ denotes the frozen FM, $C$ is the number of target classes, and $t_k$ is the predicted class probability vector.
The prediction tokens from all FMs are organized into a privileged token matrix
\begin{equation}
T=[t_1;\ldots;t_K]\in\mathbb{R}^{K\times C}.
\end{equation}

Compared with scalar performance scores, prediction tokens preserve the complete prediction distributions of FMs, providing richer instance-specific information for modeling their behaviors and supporting downstream model selection.

\subsubsection{Conditional Token Predictor.}
During inference, only the anchor model token $t_a$ is available. To
recover the unavailable FM behaviors, we introduce a
conditional token predictor.

Given the EEG representation $z_x$ and the observed anchor token $t_a$,
the predictor estimates the prediction tokens of all non-anchor
foundation models:
\begin{equation}
\hat{T}_{\setminus a}
=
P_{\psi}(z_x,t_a),
\end{equation}
\noindent
where $\hat{T}_{\setminus a}$ contains the reconstructed tokens of all
foundation models except the anchor model.

The reconstructed token matrix is defined as 
\begin{equation}
\tilde{t}_k=
\begin{cases}
t_a, & k=a,\\
\hat{t}_k, & k\neq a,
\end{cases} \text{ where } \tilde{T}
=
[\tilde{t}_1;\ldots;\tilde{t}_K].
\end{equation}
Unlike static representations shared across samples, the predicted tokens
are conditioned on the current EEG instance, enabling sample-specific
approximation of FM behaviors.

\subsubsection{Token-based Model Selector.}
Given the reconstructed token matrix $\tilde{T}$, the token-based model
selector predicts the relevance of each FM.

To model interactions between EEG characteristics and FM
behaviors, the selector employs a cross-attention mechanism \cite{vaswani2017attention}.
Each FM slot constructs a model-specific query by combining
the EEG representation with its token representation. 
These queries attend over the reconstructed token embeddings, allowing each
model slot to aggregate information from the FM portfolio.

The selector outputs relevance scores
\begin{equation}
s=f_\theta(z_x,\tilde T)\in\mathbb R^K.
\end{equation}
where $s_k$ is the predicted relevance score of the $k$-th FM.

\subsubsection{Joint Optimization.}
The conditional token predictor is first pretrained using the privileged tokens and is subsequently jointly optimized with the token-based model selector.
The token reconstruction objective is defined as
\begin{equation}
\mathcal{L}_{\mathrm{token}}
=
\frac{1}{K-1}
\sum_{k\neq a}
\|\hat t_k-t_k\|_2^2 .
\end{equation}

For an EEG instance $x$ with ground-truth label $y$, the relevance of the $k$-th FM is defined according to its prediction correctness:
\begin{equation}
r_k
=
\mathbf{1}
\left(
\arg\max_{c} t_k[c]=y
\right).
\end{equation}

Since multiple FMs may correctly predict the same instance, pairwise ranking constraints are constructed only between relevant and irrelevant models:
\begin{equation}
\mathcal{P}(x)
=
\{(i,j)\mid r_i>r_j\}.
\end{equation}

The selector is optimized using the pairwise logistic ranking loss \cite{burges2005learning}:
\begin{equation}
\mathcal{L}_{\mathrm{sel}}
=
\sum_{(i,j)\in\mathcal{P}(x)}
\log
\left(
1+\exp(-\sigma(s_i-s_j))
\right).
\end{equation}

The overall training objective is
\begin{equation}
\mathcal{L}
=
\mathcal{L}_{\mathrm{sel}}
+
\lambda
\mathcal{L}_{\mathrm{token}},
\end{equation}
\noindent
where $\lambda$ balances the token reconstruction and model selection objectives.

\subsubsection{Inference.}
At inference time, the privileged token matrix $T$ is unavailable since
obtaining it requires executing the entire FM portfolio.
Instead, EEG-AS only requires the anchor FM during selection.

Given an unseen EEG instance, we first extract its EEG representation
$z_x$ and execute the anchor model to obtain the observed token $t_a$.
The conditional token predictor then generates the missing non-anchor
tokens.
The reconstructed token matrix $\tilde{T}$ is constructed by combining
the observed anchor token with the predicted tokens. 
The token-based selector then produces relevance scores and chooses the FM with the highest score.
Therefore, EEG-AS performs efficient instance-level FM selection without exhaustive portfolio evaluation. 
If the selected model differs from the anchor model, the selected model is subsequently executed for the final downstream prediction.

\section{Experimental Setup}
\label{sec:experiment}

\subsubsection{Datasets and Evaluation Protocol.}
Following the EEG-FM-Bench evaluation protocol \cite{xiong2025eeg, xiong2026eeg}, we evaluate EEG-AS on seven publicly available EEG benchmark datasets: ADFTD \cite{miltiadous2023adftd}, BCIC-2a \cite{brunner2008bcic2a}, MIMUL-11 \cite{jeong2020mimul11}, SEED-V \cite{li2019seedv}, SEED-VII \cite{jiang2025seedvii}, THINGS-EEG-2 \cite{gifford2022thingseeg2}, Workload \cite{zyma2019workload}. These datasets covering diverse downstream tasks and recording conditions (Table~\ref{tab:datasets}).
For each dataset, we adopt the official train/validation/test split and follow the standardized downstream evaluation protocol provided by EEG-FM-Bench \cite{xiong2025eeg,xiong2026eeg}.

\begin{table}[!htbp]
\centering
{
\footnotesize
\setlength{\tabcolsep}{2pt}
\begin{tabular}{l | cccc}
\toprule
\textbf{Dataset} & Task & Classes & Subjects & Channels \\
\midrule
ADFTD        & Clinical  & 3  & 88 & 19 \\
BCIC-2a      & Motor Img.    & 4  & 9  & 22 \\
MIMUL-11     & Motor Img.   & 11 & 25 & 60 \\
SEED-V       & Emotion   & 5  & 16 & 62 \\
SEED-VII    & Emotion   & 7  & 20 & 60 \\
THINGS-EEG-2 & Visual    & 2  & 10 & 64 \\
Workload     & Cognitive & 2  & 36 & 30 \\
\bottomrule
\end{tabular}
}
\caption{Overview of the EEG benchmark datasets}
\label{tab:datasets}
\end{table}




\subsubsection{Foundation Model Portfolio.}

We construct the FM portfolio using seven representative EEG FMs, including BIOT\cite{yang2023biot}, EEGPT\cite{wang2024eegpt}, BENDR\cite{kostas2021bendr}, CBraMod\cite{wang2025cbramod}, CSBrain\cite{zhou2026csbrain}, LaBraM\cite{jiang2024labram}, and REVE\cite{el2026reve}. All FMs are frozen and are evaluated under the unified EEG-FM-Bench protocol. These models differ substantially in architecture design, pretraining strategy, and representation learning paradigm, constituting a diverse candidate portfolio for instance-level AS. 








For each EEG instance, each FM produces a class probability vector, which is used as its prediction token. These tokens provide the foundation-model behavior information required for EEG-AS training.

During inference, EEG-AS requires only one anchor model to obtain the observed prediction token. In our experiments, BIOT is selected as the anchor model and is also used to provide the EEG encoder representation. The remaining FM behaviors are reconstructed by EEG-AS without executing the full portfolio.

\subsubsection{Evaluation Metrics.}
We evaluate EEG-AS from the perspective of selection effectiveness. 
\textbf{Accuracy} is the main measure used to assess selection effectiveness:
\begin{equation}
\mathrm{Acc}_{\mathrm{sel}}
=
\frac{1}{N}
\sum_{i=1}^{N}
\rho(\hat{m}(x_i),x_i),
\end{equation}
\noindent
where $\hat{m}(x_i)$ denotes the FM selected by EEG-AS for
instance $x_i$, and $\rho(m,x)\in\{0,1\}$ indicates whether model $m$
correctly classifies instance $x$. 
This measure is bounded by the Virtual Best Solver (VBS) where the best FM is chosen for each instance.
The AS performance potential can be further realized by checking the VBS-SBS gap:
\begin{equation}
\mathrm{Gap}_{\mathrm{VBS}}
=
\mathrm{Acc}_{\mathrm{VBS}}
-
\mathrm{Acc}_{\mathrm{SBS}}.
\end{equation}
\noindent
A larger gap indicates greater instance-level diversity, offering more opportunities for AS.



\subsubsection{Baselines and Reference Methods.}
In addition to the performances of SBS and VBS / Oracle, two traditional AS models are built by using Multi-layer Perceptron (MLP) and Random Forest (RF), utilizing $z_x$. Additionally, we incorporate two reference scenarios for evaluation including:
\begin{itemize}[leftmargin=*]
    \item[--] \textbf{Global Token Predictor}
This baseline represents unavailable foundation-model outputs using
learnable global token representations. Unlike EEG-AS, the predicted tokens are independent of EEG instances.

    \item[--] \textbf{Privileged Teacher.}
Privileged Teacher receives the complete prediction token matrix generated
by all foundation models and adopts the same token-based selection
mechanism as EEG-AS. Since complete token access is unavailable during
deployment, Teacher provides a privileged upper bound rather than a
practical inference-time method.
\end{itemize}

\subsubsection{Implementation Details.}

All experiments are implemented in PyTorch. The seven EEG foundation models (FMs) remain frozen throughout training and evaluation. The EEG representation $z_x$, including the BIOT embedding and handcrafted EEG features, is precomputed and kept fixed. We choose BIOT as the default anchor FM because it provides a lightweight and practical deployment choice. Compared with other EEG foundation models, BIOT has a relatively small parameter size (3.4M parameters), making it suitable as an inference-time anchor while maintaining competitive selection performance. Therefore, EEG-AS only optimizes the conditional token predictor and the token-based selector.

The conditional token predictor and token-based selector are implemented with lightweight neural architectures. EEG-AS is optimized through the aforementioned hybridized loss $\mathcal{L}$ with $\lambda=1.0$. 
All trainable components are optimized using AdamW. All experiments are repeated with three random seeds. 
The reported results are their averages.

\section{Results}
\label{sec:results}

\subsection{Overall Selection Performance}

Table~\ref{tab:main_results} summarizes the accuracy of EEG-AS on seven EEG benchmarks. EEG-AS consistently outperforms the Single Best Solver (SBS) on all datasets, improving the average performance from 53.7\% to 66.9\%. This demonstrates the effectiveness of instance-level foundation model (FM) selection over using a single globally optimal FM.

Compared with conventional feature-based selectors, EEG-AS achieves higher average accuracy than MLP (64.8\%) and Random Forest (62.7\%), and also surpasses the Global Token Predictor (65.4\%). EEG-AS achieves the best deployable performance on five out of seven datasets, with the largest improvement observed on BCIC-2a, where it exceeds the strongest deployable baseline by 4.9 percentage points. The Privileged Teacher obtains 69.0\% by accessing complete FM prediction tokens, while the Oracle reaches 83.7\%, indicating remaining room for improving FM behavior reconstruction under practical deployment constraints.

\begin{table*}
\centering

\resizebox{\textwidth}{!}{
\begin{tabular}{l | ccccccc | cc}
\toprule
\diagbox[height=0.55cm]{\textbf{Method}}{\textbf{Dataset}}
& ADFTD 
& BCIC-2a 
& MIMUL-11 
& SEED-V 
& SEED-VII 
& THINGS-EEG-2
& Workload 
& Dataset Avg.
& Instance Avg. \\
\midrule

SBS 
& REVE
& REVE
& CSBrain
& EEGPT
& EEGPT
& REVE
& CSBrain
& 
& \\
& 60.2$\pm$0.0 
& 31.9$\pm$0.0 
& 67.9$\pm$0.0 
& 32.5$\pm$0.0 
& 29.0$\pm$0.0 
& 89.1$\pm$0.0 
& 65.6$\pm$0.0 
& 53.7$\pm$0.0 
& 68.8 \\

\midrule

MLP AS (Ours) 
& 77.2$\pm$0.7 
& 50.0$\pm$3.8 
& 72.8$\pm$0.3 
& 38.9$\pm$0.5 
& 32.5$\pm$1.5 
& 90.8$\pm$0.1 
& 91.3$\pm$0.9 
& 64.8$\pm$1.1 
& 73.7 \\

RF AS (Ours)
& 75.3$\pm$0.7 
& 46.3$\pm$1.8 
& 71.0$\pm$0.9 
& 38.7$\pm$0.9 
& 32.5$\pm$2.2 
& 89.5$\pm$0.1 
& 85.8$\pm$3.4  
& 62.7$\pm$1.4 
& 72.3 \\

Global Token Predictor
& 78.2$\pm$0.7 
& 51.1$\pm$2.0 
& \textbf{74.2$\pm$0.4} 
& 39.7$\pm$0.0 
& 32.2$\pm$3.0 
& \textbf{90.9$\pm$0.1} 
& 91.3$\pm$0.9 
& 65.4$\pm$1.0 
& 74.3 \\

Privileged Teacher
& 83.5$\pm$0.0 
& 51.4$\pm$1.3 
& 78.7$\pm$0.5 
& 42.7$\pm$0.3 
& 43.5$\pm$0.8 
& 92.0$\pm$0.1 
& 90.7$\pm$1.9 
& 69.0$\pm$0.7 
& 77.7 \\

\midrule

\textbf{EEG-AS (Ours)}
& \textbf{79.5$\pm$0.7}
& \textbf{56.0$\pm$1.5}
& 73.5$\pm$0.4
& \textbf{43.0$\pm$1.2}
& \textbf{33.7$\pm$1.3}
& 90.5$\pm$0.2
& \textbf{91.8$\pm$0.0}
& \textbf{66.9$\pm$0.7}
& \textbf{74.6}
\\

\midrule

Oracle (VBS)
& 84.0$\pm$0.0 
& 80.2$\pm$0.0 
& 90.9$\pm$0.0 
& 59.6$\pm$0.0 
& 76.1$\pm$0.0 
& 96.9$\pm$0.0 
& 98.4$\pm$0.0 
& 83.7$\pm$0.0 
& 88.9 \\

\bottomrule
\end{tabular}
}
\caption{Overall AS accuracy (\%) on seven EEG benchmarks. Learning-based selectors are reported as mean$\pm$stdev. over three random seeds, while SBS and Oracle are deterministic reference methods. Dataset Avg.\ is the macro-average across datasets; Instance Avg.\ aggregates over all evaluation instances. The best deployable method is highlighted in bold.}
\label{tab:main_results}
\end{table*}

\subsection{Ablation Studies}
To better understand the proposed framework, we conduct a series of ablation studies to answer four key questions:

\noindent
(1) Do FM embeddings and handcrafted EEG features provide complementary information for AS?

\noindent
(2) Does conditioning on an anchor FM improve the reconstruction of unavailable FM behaviors?

\noindent
(3) Which supervision target is most effective for learning the reconstructed FM behaviors?

\noindent
(4) How does the choice of the anchor FM affect the AS performance?

Unless otherwise specified, all variants follow the same training protocol as EEG-AS, and results are averaged over the seven benchmark datasets.

\subsubsection{Input Representation.}
To investigate whether FM representations provide complementary information beyond handcrafted EEG features, we compare EEG-AS under three input configurations: removing handcrafted features (\textbf{w/o Handcrafted Features}), removing the foundation-model embedding (\textbf{w/o Encoder Embedding}), and using both representations (\textbf{EEG-AS}).

As summarized Table~\ref{tab:eeg_feature_ablation}, removing the FM embedding decreases the accuracy from 66.9\% to 66.3\%, with a more noticeable degradation on BCIC-2a (56.0\% to 48.6\%). This suggests that pretrained EEG representations provide useful information for capturing instance-level characteristics that are difficult to describe using handcrafted features alone.

Meanwhile, removing handcrafted features leads to a larger overall drop when using only the encoder embedding (64.4\%). The degradation is more pronounced on SEED-V (43.0\% to 39.6\%) and Workload (91.8\% to 89.6\%), indicating that handcrafted EEG descriptors still capture complementary task-specific information. By combining both representations, EEG-AS achieves the best average performance (66.9\%), demonstrating that FM embeddings and handcrafted features provide complementary signals for FM selection.

\begin{table*}[!htbp]
\centering
\resizebox{\textwidth}{!}{%
\setlength{\tabcolsep}{2pt}
\begin{tabular}{l |ccccccc|cc}
\toprule
\textbf{Method} 
& ADFTD 
& BCIC-2a 
& MIMUL-11 
& SEED-V 
& SEED-VII 
& THINGS-EEG-2 
& Workload 
& Dataset Avg.
& Instance Avg.
\\
\midrule

w/o Handcrafted Features
& 77.6$\pm$0.7
& 48.3$\pm$1.5
& 72.9$\pm$0.9
& 39.6$\pm$0.2
& 32.5$\pm$0.8
& 90.5$\pm$0.0
& 89.6$\pm$0.9
& 64.4$\pm$0.7
& 73.6
\\

w/o Encoder Embedding
& 78.6$\pm$0.9
& 48.6$\pm$1.3
& \textbf{74.4$\pm$0.6}
& \textbf{44.0$\pm$0.7}
& \textbf{36.5$\pm$0.4}
& \textbf{90.7$\pm$0.2}
& 91.3$\pm$1.9
& 66.3$\pm$0.9
& \textbf{75.0}
\\

\textbf{EEG-AS (Ours)}
& \textbf{79.5$\pm$0.7}
& \textbf{56.0$\pm$1.5}
& 73.5$\pm$0.4
& 43.0$\pm$1.2
& 33.7$\pm$1.3
& 90.5$\pm$0.2
& \textbf{91.8$\pm$0.0}
& \textbf{66.9$\pm$0.7}
& 74.6
\\
\bottomrule
\end{tabular}%
}
\caption{Ablation study on EEG representations in EEG-AS. 
w/o Handcrafted Features removes handcrafted EEG descriptors, while 
w/o Encoder Embedding removes the FM embedding. EEG-AS uses 
both representations as the default configuration. Selection accuracy (\%) 
is mean$\pm$std over seeds.}
\label{tab:eeg_feature_ablation}
\end{table*}

\subsubsection{Anchor Conditioning}

To evaluate the importance of instance-specific anchor conditioning, we compare EEG-AS with two variants: removing the anchor token (\textbf{w/o Anchor Token}) and replacing it with a learnable global token (\textbf{Global Token Predictor}). All variants use the same EEG representation and token selector.

Table~\ref{tab:cond_ablation} shows that removing the anchor token decreases the average selection accuracy from 66.9\% to 66.3\%, while replacing it with a global token further reduces the performance to 65.9\%. The improvement is particularly evident on BCIC-2a and SEED-V, where instance-specific anchor conditioning provides more effective guidance for distinguishing FM behaviors. Although some individual datasets show comparable or slightly lower performance, EEG-AS achieves the best overall performance across the benchmark suite. 

These results indicate that EEG representations alone or dataset-level priors are insufficient to capture instance-dependent FM behaviors that an inference-available anchor FM provides valuable instance-level contextual information for reconstructing unavailable foundation-model behaviors, thereby improving downstream foundation-model selection.

\begin{table*}[t]
\centering
\resizebox{\textwidth}{!}{
\setlength{\tabcolsep}{2pt}
\begin{tabular}{l | ccccccc|cc}
\toprule
\diagbox[height=0.55cm]{\textbf{Method}}{\textbf{Dataset}}
& ADFTD 
& BCIC-2a 
& MIMUL-11 
& SEED-V 
& SEED-VII 
& THINGS-EEG-2
& Workload 
& Dataset Avg.
& Inst. Avg.
\\
\midrule

w/o Anchor Token
& 77.8$\pm$0.5
& 54.3$\pm$3.0
& 72.5$\pm$0.5
& 41.9$\pm$1.8
& \textbf{35.0$\pm$1.7}
& \textbf{90.6$\pm$0.1}
& 91.8$\pm$1.6
& 66.3$\pm$1.3
& 74.3
\\

Global Token Predictor
& 78.6$\pm$0.9
& 54.3$\pm$1.7
& 72.9$\pm$0.7
& 40.2$\pm$1.0
& 33.6$\pm$1.9
& \textbf{90.6$\pm$0.1}
& 91.3$\pm$0.9
& 65.9$\pm$1.0
& 74.1
\\

\textbf{EEG-AS (Ours)}
& \textbf{79.5$\pm$0.7}
& \textbf{56.0$\pm$1.5}
& \textbf{73.5$\pm$0.4}
& \textbf{43.0$\pm$1.2}
& 33.7$\pm$1.3
& 90.5$\pm$0.2
& \textbf{91.8$\pm$0.0}
& \textbf{66.9$\pm$0.7}
& 74.6

\\

\bottomrule
\end{tabular}}
\caption{Ablation study on the effectiveness of instance-specific anchor conditioning. w/o Anchor Token and Global Token Predictor reconstruct all foundation-model tokens without the true BIOT token, while EEG-AS conditions token reconstruction on the instance BIOT token and retains the observed anchor token during selection.}
\label{tab:cond_ablation}
\end{table*}

\subsubsection{Supervision Target}

To investigate the impact of the reconstruction target, we compare token-level supervision with correctness supervision while keeping the rest of EEG-AS unchanged.

Table~\ref{tab:target_ablation} shows that Token Supervision achieves higher overall performance than Correctness Supervision, improving the dataset-level average from 65.4\% to 66.9\% and the instance-level average from 74.1\% to 74.6\%. The improvement is particularly noticeable on BCIC-2a and SEED-V, where reconstructing prediction tokens provides richer information for distinguishing foundation-model behaviors.

This advantage comes from preserving the complete prediction distribution of each FM, including confidence and class preference information, whereas binary correctness labels discard such fine-grained details. These results validate token-level behavior reconstruction as a more informative supervision target for instance-level FM selection.
\begin{table*}[t]
\centering
\resizebox{\textwidth}{!}{
\setlength{\tabcolsep}{2pt}
\begin{tabular}{l | ccccccc|cc}
\toprule
\diagbox[height=0.55cm]{\textbf{Method}}{\textbf{Dataset}}
& ADFTD 
& BCIC-2a 
& MIMUL-11 
& SEED-V 
& SEED-VII 
& THINGS-EEG-2
& Workload 
& Dataset Avg.
& Inst. Avg.
\\
\midrule

Correctness Supervision
& \textbf{79.5$\pm$0.9}
& 49.7$\pm$1.0
& 72.7$\pm$0.2
& 39.8$\pm$1.7
& \textbf{34.5$\pm$0.4}
& \textbf{90.6$\pm$0.1}
& 90.7$\pm$0.9
& 65.4$\pm$0.7
& 74.1
\\

\textbf{Token Supervision (EEG-AS)}
& \textbf{79.5$\pm$0.7}
& \textbf{56.0$\pm$1.5}
& \textbf{73.5$\pm$0.4}
& \textbf{43.0$\pm$1.2}
& 33.7$\pm$1.3
& 90.5$\pm$0.2
& \textbf{91.8$\pm$0.0}
& \textbf{66.9$\pm$0.7}
& \textbf{74.6}
\\

\bottomrule
\end{tabular}
}
\caption{Effect of supervision target for conditional FM
behavior reconstruction. Correctness Supervision trains the predictor to
estimate FM correctness, while Token Supervision reconstructs
the full prediction tokens used in EEG-AS.}
\label{tab:target_ablation}
\end{table*}

\subsubsection{Anchor Foundation Model.}
To evaluate the robustness of EEG-AS to the choice of anchor FM, we replace the anchor FM while keeping the remaining components unchanged.
Table~\ref{tab:anchor_ablation} shows that EEG-AS maintains stable performance across different anchor choices, with dataset-level accuracy ranging from 65.7\% to 67.1\%. No single anchor consistently dominates, indicating that the proposed conditional token reconstruction mechanism can effectively exploit instance-level information from different FMs.

Although REVE achieves the highest dataset average and EEGPT obtains the highest instance average, BIOT provides comparable performance and serves as a practical default anchor because it simultaneously provides the EEG representation and anchor token required by EEG-AS.

The unified deployment results in Table~\ref{tab:unified_anchor} further confirm this observation: replacing BIOT with REVE leads to comparable selection accuracy (66.9\% vs. 67.0\%), demonstrating that EEG-AS is robust to different anchor configurations.

\begin{table*}[!htbp]
\centering
\resizebox{\textwidth}{!}{%
\setlength{\tabcolsep}{2pt}
\begin{tabular}{l | ccccccc|c| c}
\toprule
\diagbox[height=0.55cm]{\textbf{Anchor}}{\textbf{Dataset}}
& ADFTD & BCIC-2a & MIMUL-11 & SEED-V & SEED-VII & THINGS-EEG-2 & Workload & Dataset Avg. & Inst. Avg. \\
\midrule
BENDR
& 79.8$\pm$0.7 &
53.2$\pm$2.8 &
74.6$\pm$0.4 &
41.8$\pm$0.5 &
33.5$\pm$2.5 &
\textbf{91.5$\pm$0.3} &
91.8$\pm$0.0 &
66.6$\pm$1.0 &
75.1 \\
CBraMod
& \textbf{80.4$\pm$1.5} &
54.9$\pm$2.2 &
73.0$\pm$0.3 &
41.9$\pm$0.4 &
32.5$\pm$1.5 &
90.9$\pm$0.3 &
89.6$\pm$2.5 &
66.2$\pm$1.2 &
74.4 \\
CSBrain
& 78.5$\pm$0.2 &
\textbf{56.0$\pm$0.9} &
\textbf{75.1$\pm$0.6} &
\textbf{42.5$\pm$0.8} &
32.9$\pm$1.9 &
90.8$\pm$0.1 &
90.2$\pm$2.8 &
66.6$\pm$1.0 &
74.9 \\
EEGPT
& 78.2$\pm$0.2 &
52.3$\pm$0.5 &
73.9$\pm$1.2 &
41.5$\pm$0.5 &
\textbf{38.6$\pm$0.9} &
91.4$\pm$0.2 &
90.7$\pm$1.9 &
66.7$\pm$0.8 &
\textbf{75.3} \\
LaBraM
& 78.2$\pm$0.9 &
51.4$\pm$1.3 &
73.3$\pm$0.7 &
41.6$\pm$1.3 &
33.8$\pm$2.6 &
90.6$\pm$0.3 &
91.3$\pm$0.9 &
65.7$\pm$1.1 &
74.2 \\
REVE
& 78.4$\pm$0.4 &
55.5$\pm$1.8 &
72.8$\pm$0.9 &
41.8$\pm$0.8 &
37.4$\pm$0.7 &
91.3$\pm$0.1 &
\textbf{92.3$\pm$1.9} &
\textbf{67.1$\pm$0.9} &
75.0 \\
\textbf{BIOT (Ours)}
& 80.2$\pm$0.7 &
54.3$\pm$0.0 &
73.2$\pm$0.4 &
42.3$\pm$0.4 &
32.9$\pm$1.3 &
90.5$\pm$0.1 &
91.3$\pm$2.5 &
66.4$\pm$0.8 &
74.3 \\
\bottomrule
\end{tabular}%
}
\caption{Robustness of EEG-AS to different anchor FMs. Each variant replaces only the inference-available anchor token while keeping the EEG representation, conditional token predictor, token-based selector, and training protocol unchanged.}
\label{tab:anchor_ablation}
\end{table*}

\begin{table*}[!htbp]
\centering
\resizebox{\textwidth}{!}{%
\setlength{\tabcolsep}{2pt}
\begin{tabular}{l |  ccccccc|c| c}
\toprule
\diagbox[height=0.55cm]{\textbf{Anchor}}{\textbf{Dataset}}
& ADFTD 
& BCIC-2a 
& MIMUL-11 
& SEED-V 
& SEED-VII 
& THINGS-EEG-2 
& Workload 
& Dataset Avg.
& Inst. Avg.
\\
\midrule

BIOT
& \textbf{79.5$\pm$0.7}
& \textbf{56.0$\pm$1.5}
& 73.5$\pm$0.4
& 43.0$\pm$1.2
& 33.7$\pm$1.3
& 90.5$\pm$0.2
& \textbf{91.8$\pm$0.0}
& 66.9$\pm$0.7
& 74.6
\\

REVE
& 77.9$\pm$0.0
& 48.9$\pm$2.0
& \textbf{78.4$\pm$0.4}
& \textbf{44.3$\pm$0.7}
& \textbf{37.5$\pm$1.3}
& \textbf{91.2$\pm$0.1}
& 90.7$\pm$1.9
& \textbf{67.0$\pm$0.9}
& \textbf{76.3}
\\

\bottomrule
\end{tabular}%
}
\caption{
Unified deployment with different anchor foundation models.
Each variant uses the same foundation model for both EEG representation
extraction and anchor token generation.
}
\label{tab:unified_anchor}
\end{table*}

\subsection{Analysis}

\subsubsection{Reconstructed Foundation-model Behavior.}
\begin{figure}[!t]
\centering
\includegraphics[width=\linewidth]{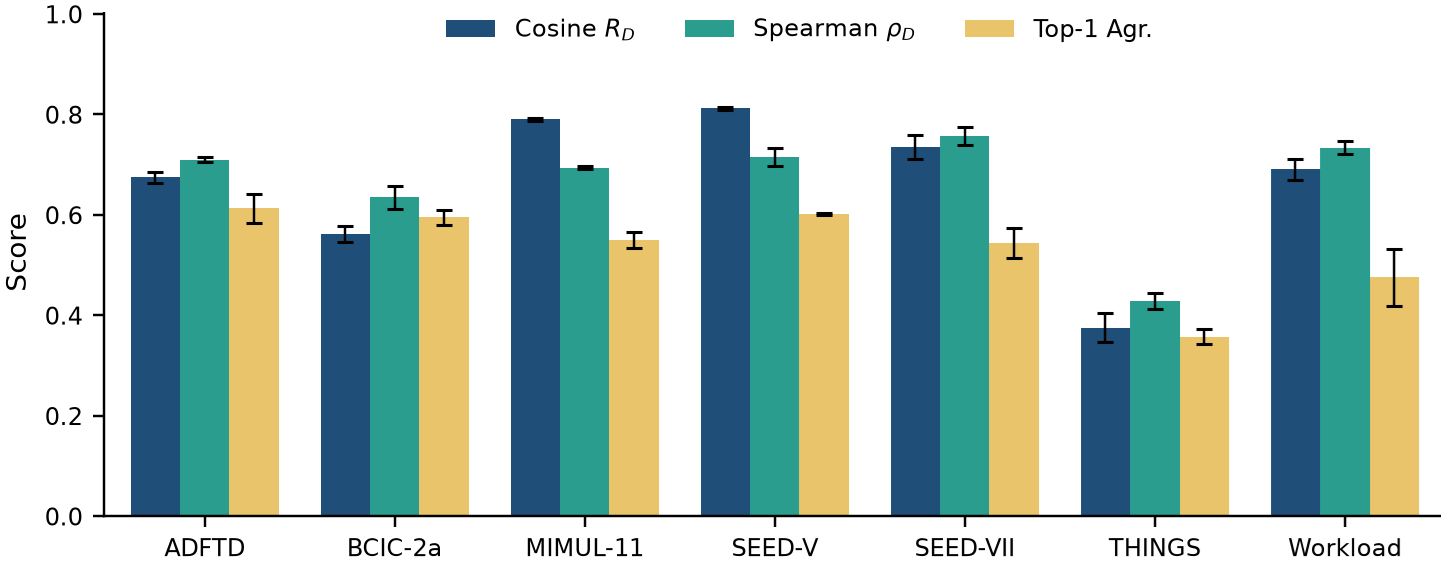}
\caption{
Token-reconstruction metrics on each dataset (mean$\pm$std over seeds: cosine similarity $R_D$, Spearman rank correlation $\rho_D$, and Top-1 agreement of $\arg\max_m P(y\mid z)$.
}
\label{fig:token_reconstruction_bars}
\end{figure}

To verify whether EEG-AS can recover unavailable foundation-model (FM)  behaviors from an inference-available anchor FM, we evaluate the reconstructed 
tokens using cosine similarity, Spearman rank correlation between FM rankings, and Top-1 agreement between the reconstructed and true best FMs.
Table~\ref{tab:token_reconstruction} and Fig.~\ref{fig:token_reconstruction_bars} summarize the results.
Overall, EEG-AS achieves an average cosine similarity of 0.66, a ranking correlation of 0.67, and a Top-1 agreement of 0.53, demonstrating that the conditional token predictor captures meaningful FM behaviors rather than learning arbitrary representations.

The reconstruction quality varies across datasets. 
SEED-V and MIMUL-11 show the highest reconstruction fidelity, while THINGS-EEG-2 remains more challenging as the relationships among FM behaviors are dataset dependent. 

Moreover, Top-1 agreement is consistently lower than token similarity and ranking correlation, suggesting that reconstructing the general behavior of FMs is easier than precisely identifying the optimal FM for each instance.

\begin{table}[!t]
\centering
{\setlength{\tabcolsep}{2pt}
\begin{tabular}{l | ccc}
\toprule
\textbf{Dataset} & $R_D$ & Ranking $\rho_D$ & Top-1 Agr. \\
\midrule
ADFTD & 0.67$\pm$0.01 & 0.71$\pm$0.01 & 0.61$\pm$0.03 \\
BCIC-2a & 0.56$\pm$0.02 & 0.63$\pm$0.02 & 0.59$\pm$0.01 \\
MIMUL-11 & 0.79$\pm$0.00 & 0.69$\pm$0.00 & 0.55$\pm$0.02 \\
SEED-V & 0.81$\pm$0.00 & 0.71$\pm$0.02 & 0.60$\pm$0.00 \\
SEED-VII & 0.74$\pm$0.02 & 0.76$\pm$0.02 & 0.54$\pm$0.03 \\
THINGS-EEG-2 & 0.38$\pm$0.03 & 0.43$\pm$0.02 & 0.36$\pm$0.01 \\
Workload & 0.69$\pm$0.02 & 0.73$\pm$0.01 & 0.48$\pm$0.06 \\
\midrule
Average & 0.66 & 0.67 & 0.53 \\
\bottomrule
\end{tabular}
}
\caption{Token reconstruction analysis. $R_D$: mean cosine similarity between predicted and true tokens over unavailable FMs. $\rho_D$: mean per-instance Spearman correlation of FM rankings by $P(y\mid z)$. Top-1: agreement of the highest-$P(y)$ model under true vs.\ reconstructed tokens ($=1/|\mathcal{M}|$).}
\label{tab:token_reconstruction}
\end{table}

\subsubsection{Understanding Performance Gaps.}
Although EEG-AS effectively reconstructs unavailable FM behaviors, the 
selection gains vary across datasets. We identify three factors that limit 
the final AS performance.

First, the achievable improvement is bounded by the selection headroom 
between SBS and Oracle. For example, on THINGS-EEG-2, SBS already achieves 
89.1\% accuracy while Oracle reaches 96.9\%, leaving limited room for 
instance-level selection improvements. Therefore, the modest gain on this 
dataset is mainly due to the limited selection margin rather than ineffective 
behavior reconstruction.

Second, accurate token reconstruction does not necessarily guarantee optimal 
FM selection. On SEED-VII, EEG-AS achieves strong reconstruction quality 
($R_D=0.74$, $\rho_D=0.76$), but the selection accuracy remains limited. 
Moreover, the Privileged Teacher using true FM tokens only achieves 43.5\%, 
far below the Oracle performance of 76.1\%. These results reveal that FM selection is not solely a representation reconstruction problem, but a decision alignment problem.

Finally, FM diversity affects the effectiveness of selection. When multiple 
FMs exhibit highly similar prediction behaviors, their tokens provide limited 
discriminative information for identifying the optimal model.

Overall, these findings suggest that effective FM selection requires not only 
accurate behavior reconstruction, but also sufficient selection headroom and 
decision-relevant diversity among candidate FMs.

\section{Conclusion}
\label{sec:conc}

In this paper, we propose EEG-AS, an instance-level Algorithm Selection (AS)
framework for EEG foundation models (FMs). 
By reconstructing unavailable
FM behaviors from an inference-available anchor model, EEG-AS enables adaptive selection of FMs for different EEG instances.

Experiments on seven EEG benchmarks demonstrate that EEG-AS consistently
outperforms conventional selection strategies as well as standalone FMs. 
Further analysis shows that the reconstructed tokens capture meaningful FM behaviors,
while the remaining gap depends on the difficulty of identifying decision-relevant model behaviors.

In the future, we will investigate cost-aware FM selection by jointly considering model accuracy, inference time, and computational resources.


\bibliography{aaai2027}


\end{document}